\documentclass{article}

\usepackage[dblblindworkshop,final]{neurips_2026}
\workshoptitle{AI for Drug Discovery -- Bridging the Translation Gap}
\usepackage{amsmath}
\usepackage{amssymb}
\usepackage{graphicx}
\usepackage{booktabs}
\usepackage{xcolor}
\usepackage{hyperref}

\title{Molecular Prior Networks}
\author{
    Pawel Gomoluch$^{1}$, Anthony Bradley$^{1,2}$ \\
    $^{1}$DaltonTx \quad $^{2}$Department of Chemistry, University of Liverpool \\
    \texttt{pgomoluch@daltontx.com}, \texttt{abradley@daltontx.com}
}

\begin{document}

\maketitle

\begin{abstract}

    Predicting affinity between small molecules and their protein targets remains one of the central problems in drug discovery.
    Due to the substantial cost of wet-lab experiments, machine learning models of drug-target affinity are usually trained on relatively small datasets, ranging from tens to hundreds of molecules.
    Given the inherent difficulty of generalization between different targets and chemical series,
    existing datasets of protein-ligand affinities are typically not used for new projects.
    At the same time, tabular foundation models, including Prior-data Fitted Networks, have recently shown how transformers can be used to solve machine learning problems in a single forward pass of the network,
    after being trained over a large corpus of, typically, synthetic datasets.
    Inspired by that line of work, we propose a novel affinity modelling approach, in which the labels of the query molecules are predicted from the labelled support molecules in a single forward pass of a transformer model.
    The model is trained on thousands of existing affinity datasets, enabling it to capture the general patterns of molecular activity across different targets, chemical series and dataset sizes.
    For the smallest support sets, our model matches the state of the art on the regression variant of the FS-Mol benchmark, while being faster than existing methods, thanks to the lack of task-specific training.
    A comparison against a reduced variant of our architecture, which amounts to kernel regression in a meta-learned feature space, shows that the predictive performance of the transformer model is only partially explained by simple molecular similarity.

\end{abstract}

\section{Introduction}

    Recent developments in tabular foundation models have demonstrated a remarkable ability of deep neural networks to generalize between tabular datasets and perform inference on unseen datasets in a single forward pass.
    Models such as TabPFN \citep{Hollmann2023,Grinsztajn2025} and TabICL \citep{Qu2025}, trained over thousands of synthetic datasets, have achieved state-of-the-art performance on real-world tabular machine learning problems.
    In this work, we propose using a similar transformer-based approach to build drug-target affinity models.
    Instead of synthetically generated datasets, we use the wealth of existing affinity datasets to imbue our model with the relevant priors. 
    Over the course of training, the datasets are randomly split into support and query sets and the model is tasked with predicting the labels of the query molecules, using the information contained in the support molecules and their labels.
    At inference, the model is presented with an unseen dataset including both labelled and unlabelled examples, and predicts the missing labels in a single forward pass, without the need for any task-specific training.

    Since Prior-data Fitted Networks have been hypothesized to approximate non-parametric inference \citep{Muller2022}, we also introduce a reduced variant of our approach, in which similarity between projected molecule representations is used directly to compute a weighted mean of the support labels. While this model outperforms common baselines, it remains behind the full transformer model.

    The source code of our models and experiments and the trained model weights are publicly available from \url{https://github.com/daltontx/molpriornet}.

\section{Related work}

    By using a large collection of molecular property datasets to improve predictive performance on unseen datasets, our work naturally falls within the field of molecular meta-learning.
    In our experiments, we show that our model matches state-of-the-art predictive performance on the 16-shot regression tasks from the FS-Mol benchmark \citep{Stanley2021},
    while saving computation time due to the lack of task-specific training.

    The general idea of using attention between labelled and unlabelled molecules is not entirely new. Restricted forms of attention have previously been explored as part of a number of more complex architectures.
    For instance, the sum of support labels weighted by the similarity of learned embeddings is the output mechanism for the \emph{iterative refinement LSTMs} in \citet{Altae-Tran2017}.
    An attention layer with values representing the support labels is also a key component of the Attentive Neural Process \citep{Kim2019}, applied to molecular property prediction by \citet{Lee2022}.
    
    A single layer of attention between query and support molecules forms the cross-attention module of MHNfs \citep{Schimunek2023,Schimunek2025}. It is then followed by a similarity module, which uses molecule similarity to compute a weighted mean of the support labels for each query molecule, in a manner similar to our kernel baseline. Together with the \emph{context module}, in which a Modern Hopfield Network is used to retrieve related molecules from a subsample of the FS-Mol dataset, this architecture leads to very strong performance in the 16-shot classification tasks of the FS-Mol benchmark.
    Unlike MHNfs, our main model processes features and labels jointly at every layer and does not use the similarities to directly average support labels. This results in a conceptually simpler architecture, while also allowing for stacking multiple transformer layers, which we find essential for predictive performance.
    Furthermore, our model does not rely on external context beyond the information stored in the weights of the network. This is especially important for inference on larger datasets, as the cost of attending to a large context scales linearly with the sizes of the support and query sets.
    Our work is also generally geared towards regression problems, since we are primarily interested in accurate ranking of candidate molecules.

    In bridging parametric and non-parametric modelling, our work is also related to deep-kernel Gaussian processes, and in particular to ADKF-IFT \citep{Chen2023}, the state of the art in few-shot molecular regression. ADKF-IFT combines a meta-trained graph neural network (GNN) encoder with a Gaussian process fit to individual tasks and generally outperforms other approaches to data-efficient adaptation of GNNs. The optimization scheme based on the Implicit Function Theorem enables efficient meta-learning despite the nested optimization setting.

\section{Approach}

\subsection{Molecular Prior Network}

    The Molecular Prior Network (MPN) is a transformer that takes as input a set of molecules, partitioned into a labelled \emph{support} set
    $\mathcal{S} = \{(x_i^s, y_i^s)\}_{i=1}^{N_s}$ and an unlabelled \emph{query} set $\mathcal{Q} = \{x_j^q\}_{j=1}^{N_q}$,
    and jointly predicts the labels $\hat y_j^q$ of all query molecules in a single forward pass.
    Each molecule is represented by a fixed-length binary fingerprint $x \in \{0,1\}^F$; labels $y \in \mathbb{R}$ are real-valued regression targets.
    Unlike TabPFN, where individual features are tokenized, we treat each molecule as a single token and keep the feature ordering fixed across datasets.
    This lets the model operate under fixed semantics of individual fingerprint bits and enables learning molecular patterns that generalize across targets.

    Following the original TabPFN architecture, attention is asymmetric and only the support molecules are attended to, so that evidence flows only from the labelled molecules. 
    Otherwise, the support and query molecules are processed by the same stack of layers. The predictions $\hat y_j^q$ are read off the hidden states of the query molecules at the final layer by a single linear head.
    While the support hidden states are not used for the output, they are repeatedly attended to and updated, allowing the network to construct task-specific support representations conditioned on the labels.

    \paragraph{Input embedding.}
    Molecules are embedded into a $d$-dimensional token space by a shared linear map, while support labels are normalized by subtracting the task support mean and dividing by the standard deviation, and embedded by a separate linear map into a $d_y$-dimensional space (we use $d_y = d/4$):
    \begin{equation}
        h_i^{(0)} = W_h\, x_i + b_h, \quad i \in \{1, \dots, N_s + N_q\}, \qquad
        e_i^{y} = W_y\, y_i^s, \quad i \in \{1, \dots, N_s\},
    \end{equation}
    where the first $N_s$ tokens correspond to support molecules and the remaining $N_q$ to queries.
    No positional encoding is added, so the model predictions are invariant to permutations of the support set.

    \paragraph{Transformer block.}
    The stack consists of $L$ residual blocks. Within block $\ell$ we first form \emph{value} tokens by concatenating the current support hidden states with the label embeddings,
    \begin{equation}
        v_i^{(\ell)} = \big[\, h_i^{(\ell)} \,;\, e_i^{y} \,\big] \in \mathbb{R}^{d + d_y}, \qquad i \in \{1, \dots, N_s\}.
        \label{eq:values}
    \end{equation}
    Attention is then applied with pre-normalization, $\tilde h^{(\ell)} = \mathrm{LayerNorm}(h^{(\ell)})$, and a learned temperature $\tau$:
    \begin{equation}
        h^{(\ell + \tfrac12)} = h^{(\ell)} + \mathrm{Attn}\!\left( Q = \tilde h^{(\ell)}/\tau,\; K = \tilde h^{(\ell)}_{1:N_s},\; V = v^{(\ell)} \right),
        \label{eq:attn}
    \end{equation}
    where $\mathrm{Attn}$ is single-head scaled dot-product attention with learned projections $W_Q, W_K, W_V$ and output projection $W_O$:
    \begin{equation}
        \mathrm{Attn}(Q, K, V)_i = W_O \sum_{j=1}^{N_s} \alpha_{ij}\, W_V v_j, \qquad
        \alpha_{ij} = \frac{\exp\!\big(\langle W_Q q_i,\, W_K k_j \rangle / \sqrt{d}\big)}{\sum_{j'=1}^{N_s} \exp\!\big(\langle W_Q q_i,\, W_K k_{j'} \rangle / \sqrt{d}\big)}.
    \end{equation}
    Note that labels enter the computation exclusively through the values of attention, via the concatenation in Eq.~\eqref{eq:values}.
    The original label embeddings are re-injected at every layer, which we have found to improve the model's performance.

    In line with modern transformer architectures \citep{Karpathy2025}, the attention is followed by a small feed-forward network with squared ReLU activation and a residual connection,
    \begin{equation}
        h^{(\ell + 1)} = h^{(\ell + \tfrac12)} + W_2 \Big[\,\mathrm{ReLU}\!\big( W_1\, \mathrm{LayerNorm}(h^{(\ell + \tfrac12)}) \big)\,\Big]^{\!2},
    \end{equation}
    where the inner dimension of the FFN is $4d$.

    \paragraph{Output head.}
    After $L$ blocks, predictions are read off the query tokens by a linear head,
    \begin{equation}
        \hat y_j^q = W_o\, h^{(L)}_{N_s + j}, \qquad j \in \{1, \dots, N_q\}.
    \end{equation}

\subsection{Kernel baseline} \label{sec:kernel}

    We hypothesize that the MPN may be effectively implementing a form of non-parametric inference, determining the query labels as weighted sums of the labels of the nearest support molecules.
    To test this, we introduce a reduced architecture which makes this structure explicit.
    The resulting \emph{kernel MPN} discards the transformer stack, and instead predicts each query label as a similarity-weighted average of the support labels, using a single meta-learned feature projection.

    Given the same support set $\mathcal{S} = \{(x_i^s, y_i^s)\}_{i=1}^{N_s}$ and query set $\mathcal{Q} = \{x_j^q\}_{j=1}^{N_q}$ as before, the kernel baseline embeds all molecules with a shared linear projection $\phi: \mathbb{R}^F \to \mathbb{R}^d$,
    and $L^2$-normalizes the result to the unit sphere.
    Similarity between a query and a support molecule is measured by the cosine of the angle between their normalized projections, scaled by a learned temperature $\tau$:
    \begin{equation}
        s_{ji} = \frac{\phi(x_j^q)^\top \phi(x_i^s)}{\tau}, \quad \text{where} \quad \phi(x) = \frac{W_\phi\, x}{\lVert W_\phi\, x \rVert_2}.
    \end{equation}
    Predictions are a softmax-weighted mean of the support labels,
    \begin{equation}
        \hat y_j^q = \sum_{i=1}^{N_s} \alpha_{ji}\, y_i^s, \qquad
        \alpha_{ji} = \frac{\exp(s_{ji})}{\sum_{i'=1}^{N_s} \exp(s_{ji'})},
        \label{eq:kernel-pred}
    \end{equation}
    which makes the model an instance of kernel regression with an exponential-cosine kernel and a meta-learned feature projection $\phi$.

\subsection{Training} \label{sec:training}

    Both of the architectures described above are trained end-to-end by meta-learning over a corpus of existing affinity datasets $\{\mathcal{D}_k\}_{k=1}^{K}$.
    Each dataset $\mathcal{D}_k = \{(x_i, y_i)\}_{i=1}^{n_k}$ corresponds to a single protein target and comprises pairs of molecular fingerprints and real-valued activity measurements.
    We denote the full set of learnable parameters by $\theta$ and the network's prediction on query $x_j^q$ given support set $\mathcal{S}$ by $\hat y_j^q(\mathcal{S}; \theta)$.

    At each training step, a dataset $\mathcal{D}$ is sampled from the corpus and randomly partitioned into a support set $\mathcal{S}$ and a query set $\mathcal{Q}$ of sizes drawn from a distribution over $(N_s, N_q)$.
    The parameters are updated to minimize the mean squared error of the predicted query labels,
    \begin{equation}
        \mathcal{L}(\theta) = \mathbb{E}_{\mathcal{D}}\, \mathbb{E}_{(\mathcal{S}, \mathcal{Q}) \sim \mathcal{D}}\,
        \frac{1}{N_q} \sum_{j=1}^{N_q} \big( \hat y_j^q(\mathcal{S}; \theta) - y_j^q \big)^2.
        \label{eq:loss}
    \end{equation}

\section{Evaluation} \label{sec:evaluation}

    \begin{table}
        \centering
        \caption{Kendall's $\tau$ on the FS-Mol test tasks as a function of the support set size $N_s$. The $N_s = 256$ column is computed on the subset of tasks (43 out of 111) with enough samples to form a $256$-molecule support set.}
        \label{tab:kendall}
        \footnotesize
        \begin{tabular}{lccccc}
            \toprule
            & \multicolumn{5}{c}{Support set size $N_s$} \\
            \cmidrule(lr){2-6}
            Model           & 16              & 32              & 64              & 128             & 256             \\
            \midrule
            Random forest   & 0.138           & 0.198           & 0.255           & 0.310           & 0.354           \\
            GP-ST           & 0.163           & 0.214           & 0.262           & 0.315           & 0.358           \\
            ADKF-IFT        & 0.298           & 0.349           & 0.391           & 0.434           & 0.456           \\
            TabPFN          & 0.146           & 0.190           & 0.228           & 0.276           & 0.322           \\
            TabICL (500)    & 0.150           & 0.197           & 0.247           & 0.304           & 0.350           \\
            TabICL (2048)   & 0.156           & 0.200           & 0.250           & 0.309           & 0.358           \\
            kernel MPN      & 0.233           & 0.282           & 0.326           & 0.369           & 0.379           \\
            MPN (4 layers)  & 0.299           & 0.326           & 0.352           & 0.382           & 0.395           \\
            MPN (16 layers) & 0.298           & 0.335           & 0.368           & 0.398           & 0.411           \\
            \bottomrule
        \end{tabular}
    \end{table}

\subsection{Experimental setup}

    We evaluate the MPN models on the regression variant of the FS-Mol benchmark \citep{Stanley2021,Chen2023}. We train on the 4938 training tasks, using the logarithmic regression targets (i.e., the pIC50 values). To avoid spending a disproportionate amount of time on a small group of datasets with a large number of samples, we randomly subsample sets larger than 500 molecules to 500 samples. The support size is uniformly sampled from the set $\{16, 32, 64, 128, 256\}$ and capped at 80\% of the entire dataset, so that at least 20\% of the dataset is used as the query. Subsampling and splitting of individual datasets is performed independently at each epoch.
    We use the mean absolute error computed on the 40 validation tasks to select the best model within the training run,
    and report evaluation metrics averaged across the 111 test tasks and 10 different random splits for each task and support set size.

    The molecules are featurized with Morgan fingerprints of radius 2 and length 2048.
    The attention-based models use a hidden dimension of 1024 while the kernel baseline linearly projects the input without changing its dimensionality. All MPN models are trained with the Adam optimizer, with a learning rate of $10^{-4}$ for the kernel variant and $10^{-5}$ for the attention-based MPNs, batches of 10 tasks, and gradient norm clipped at 100.
    The kernel MPN is trained for a total of 100 epochs and the attention-based MPN for 400 epochs. In each case, we keep the checkpoint with the lowest validation MAE.
    The total time required to train a 16-layer MPN is approximately 4 days on a single NVIDIA A10G GPU.

    We compare our models against ADKF-IFT \citep{Chen2023}, using a model checkpoint provided by the authors, TabPFNv2 \citep{Hollmann2025} (fingerprints folded to 500 bits to match the model's limit), and TabICLv2 \citep{Qu2026} (both 500 and 2048-bit fingerprints). We also include the baselines of a tuned random forest and a Gaussian process with a Tanimoto kernel (GP-ST), fitted independently on the support set of every task, both of which we rerun without any changes from \citet{Chen2023}. We report the Kendall rank correlation coefficient $\tau$ between true and predicted affinities, which aligns with our interest in accurate ranking of the molecules.

\subsection{Results}

\begin{figure}
        \centering
        \begin{minipage}[t]{0.48\linewidth}
            \centering
            \includegraphics[width=\linewidth]{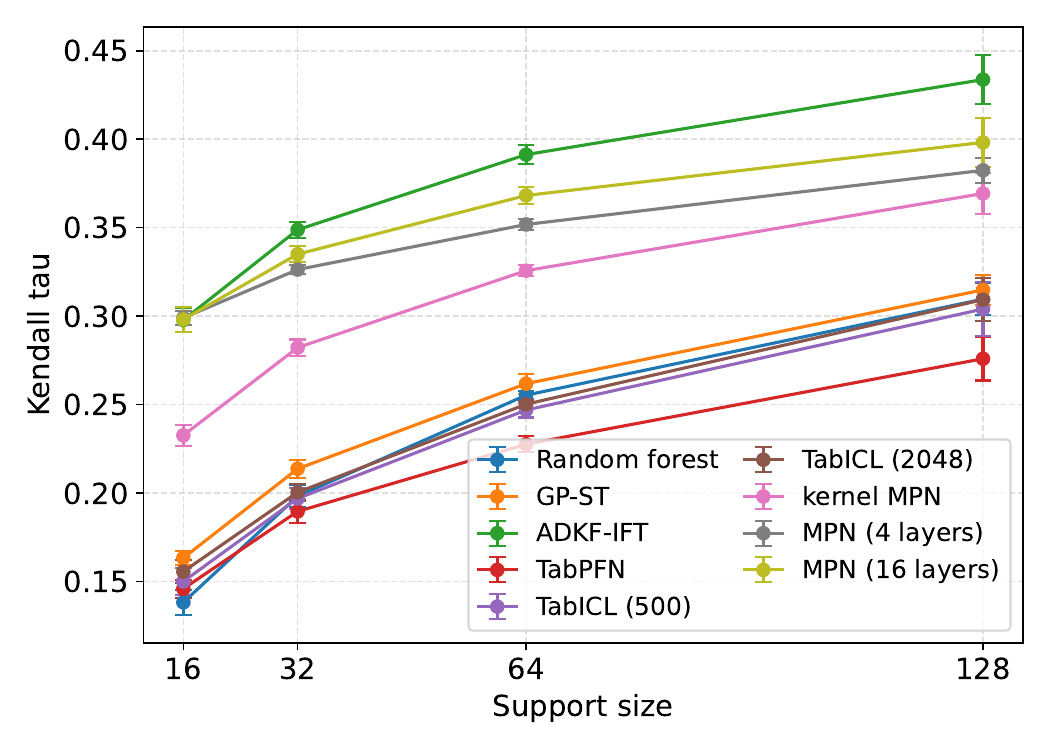}
            \caption{Kendall's $\tau$ on the FS-Mol test tasks as a function of the support set size. Error bars: standard deviation over 10 seeds.}
            \label{fig:tau}
        \end{minipage}
        \hfill
        \begin{minipage}[t]{0.48\linewidth}
            \centering
            \includegraphics[width=\linewidth]{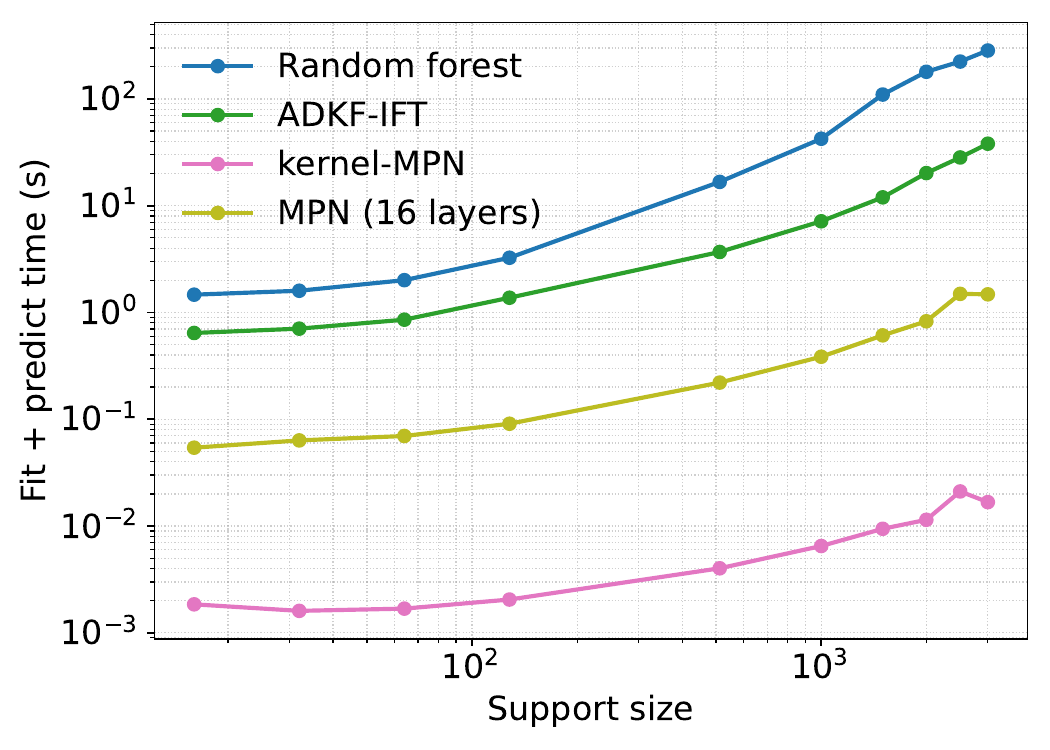}
            \caption{Joint fit and inference time as a function of the support set size (100 query molecules, 10 CPU cores, median of 3 runs).}
            \label{fig:runtime}
        \end{minipage}
    \end{figure}

    The Kendall rank correlation coefficient for test tasks is reported in Table~\ref{tab:kendall}, with the corresponding plots in Figure~\ref{fig:tau}.
    Overall, the strongest method remains ADKF-IFT, but for the smallest support set size of 16, both attention-based MPN models are able to match its performance.
    The kernel baseline turns out to be a surprisingly strong model, given that it performs no fitting on the evaluation task and reduces inference to a single matrix multiplication followed by a softmax operation and a weighted sum.
    It outperforms both random forests and basic Gaussian processes at every support set size.
    This supports the hypothesis of Section~\ref{sec:kernel}: a substantial fraction of the predictive performance achievable on FS-Mol can be expressed as kernel regression in a meta-learned feature space.
    The comparison between MPNs and general tabular foundation models shows that in settings for which a domain-specific pre-training corpus is available, it can lead to stronger performance than generic priors. 

    To illustrate the computational efficiency of MPNs, we pick one of the largest test datasets in FS-Mol and plot the joint time needed for fitting and inference on a fixed query set of 100 samples, as a function of the support set size.
    The results are shown in Figure~\ref{fig:runtime}. MPNs are capable of performing inference over thousands of molecules in a matter of seconds (attention) or hundredths of a second (kernel), which should allow future work to apply the same architecture across a wider range of support set sizes.
    We remark that both MPN and ADKF-IFT are roughly two times faster on a GPU, but GPU memory becomes a bottleneck for ADKF-IFT beyond 1000 samples.

\section{Conclusion}

    In this work, we introduce the Molecular Prior Network (MPN), a new approach to protein-ligand affinity modelling.
    Inspired by Prior-data Fitted Networks, the MPN jointly processes labelled support and unlabelled query molecules given as inputs and predicts the labels of the query molecules, all in a single forward pass of the transformer model.
    
    The MPN architecture opens a number of future research directions. One of them is explicit modelling of prediction uncertainty, which would enable the use of the model in active learning and Bayesian optimization settings. Another possibility is going beyond experimental affinity values, and training on a wider range of both experimentally and computationally derived molecular properties, turning MPNs into a foundation model for molecular property prediction. An orthogonal question is the representation of the input molecules: Morgan fingerprints could be augmented with additional descriptors or replaced with a graph neural network encoder, trained end-to-end with the transformer layers.

\bibliographystyle{abbrvnat}
\bibliography{references}

@article{Altae-Tran2017,
  title={Low data drug discovery with one-shot learning},
  author={Altae-Tran, Han and Ramsundar, Bharath and Pappu, Aneesh S and Pande, Vijay},
  journal={ACS central science},
  volume={3},
  number={4},
  pages={283--293},
  year={2017},
  publisher={ACS Publications}
}

@inproceedings{Chen2023,
  title={Meta-learning Adaptive Deep Kernel {G}aussian Processes for Molecular Property Prediction},
  author={Chen, Wenlin and Tripp, Austin and Hern{\'a}ndez-Lobato, Jos{\'e} Miguel},
  booktitle={The Eleventh International Conference on Learning Representations},
  year={2023}
}

@article{Grinsztajn2025,
  title={{TabPFN}-2.5: Advancing the state of the art in tabular foundation models},
  author={Grinsztajn, L{\'e}o and Fl{\"o}ge, Klemens and Key, Oscar and Birkel, Felix and Jund, Philipp and Roof, Brendan and J{\"a}ger, Benjamin and Safaric, Dominik and Alessi, Simone and Hayler, Adrian and others},
  journal={arXiv preprint arXiv:2511.08667},
  year={2025}
}

@inproceedings{Hollmann2023,
  title={{TabPFN}: A Transformer That Solves Small Tabular Classification Problems in a Second},
  author={Hollmann, Noah and M{\"u}ller, Samuel and Eggensperger, Katharina and Hutter, Frank},
  booktitle={The Eleventh International Conference on Learning Representations},
  year={2023}
}

@article{Hollmann2025,
  title   = {Accurate predictions on small data with a tabular foundation model},
  author  = {Hollmann, Noah and M{\"u}ller, Samuel and Purucker, Lennart and Krishnakumar, Arjun and K{\"o}rfer, Max and Hoo, Shi Bin and Schirrmeister, Robin Tibor and Hutter, Frank},
  journal = {Nature},
  volume  = {637},
  number  = {8045},
  pages   = {319--326},
  year    = {2025},
  doi     = {10.1038/s41586-024-08328-6}
}

@misc{Karpathy2025,
  author = {Andrej Karpathy},
  title = {nanochat: The best {ChatGPT} that \$100 can buy},
  year = {2025},
  publisher = {GitHub},
  url = {https://github.com/karpathy/nanochat}
}

@inproceedings{Kim2019,
  title={Attentive Neural Processes},
  author={Kim, Hyunjik and Mnih, Andriy and Schwarz, Jonathan and Garnelo, Marta and Eslami, Ali and Rosenbaum, Dan and Vinyals, Oriol and Teh, Yee Whye},
  booktitle={International Conference on Learning Representations},
  year={2019}
}

@inproceedings{Lee2022,
  title={{MetaDTA}: Meta-learning-based drug-target binding affinity prediction},
  author={Lee, Eunjoo and Yoo, Jiho and Lee, Huisun and Hong, Seunghoon},
  booktitle={ICLR2022 Machine Learning for Drug Discovery},
  year={2022}
}

@inproceedings{Muller2022,
  title={Transformers Can Do {B}ayesian Inference},
  author={M{\"u}ller, Samuel and Hollmann, Noah and Arango, Sebastian Pineda and Grabocka, Josif and Hutter, Frank},
  booktitle={International Conference on Learning Representations},
  year={2022}
}

@inproceedings{Qu2025,
  title={Tab{ICL}: {A} Tabular Foundation Model for In-Context Learning on Large Data},
  author={Qu, Jingang and Holzm{\"u}ller, David and Varoquaux, Ga{\"e}l and Le Morvan, Marine},
  booktitle={International Conference on Machine Learning},
  year={2025}
}

@article{Qu2026,
  title={{TabICLv2}: {A} better, faster, scalable, and open tabular foundation model},
  author={Qu, Jingang and Holzm{\"u}ller, David and Varoquaux, Ga{\"e}l and Le Morvan, Marine},
  journal={arXiv preprint arXiv:2602.11139},
  year={2026}
}

@inproceedings{Schimunek2023,
  title={Context-enriched molecule representations improve few-shot drug discovery},
  author={Schimunek, Johannes and Seidl, Philipp and Friedrich, Lukas and Kuhn, Daniel and Rippmann, Friedrich and Hochreiter, Sepp and Klambauer, G{\"u}nter},
  booktitle={The Eleventh International Conference on Learning Representations},
  year={2023}
}

@article{Schimunek2025,
  title={{MHNfs}: Prompting In-Context Bioactivity Predictions for Low-Data Drug Discovery},
  author={Schimunek, Johannes and Luukkonen, Sohvi and Klambauer, Günter},
  journal={Journal of Chemical Information and Modeling},
  volume={65},
  number={9},
  pages={4243--4250},
  year={2025},
  publisher={ACS Publications}
}

@inproceedings{Stanley2021,
  title={{FS-M}ol: A few-shot learning dataset of molecules},
  author={Stanley, Megan and Bronskill, John F and Maziarz, Krzysztof and Misztela, Hubert and Lanini, Jessica and Segler, Marwin and Schneider, Nadine and Brockschmidt, Marc},
  booktitle={Thirty-fifth Conference on Neural Information Processing Systems Datasets and Benchmarks Track (Round 2)},
  year={2021}
}

\end{document}